# Improved YOLOv5 network for real-time multi-scale traffic sign detection

Junfan WANG[1], Yi CHEN[1], Mingyu GAO (✉) [1,2], Zhenkang DONG[1,3]

1  School of Electronic Information, Hangzhou Dianzi University, Hangzhou 310018, China
2  Zhenjiang Provincial Key Lab of Equipment Electronics, Hangzhou 310018, China
3  Department of Electronic Engineering, Zhejiang University, Hangzhou 310027, China

**Abstract** Traffic sign detection is a challenging task for the unmanned driving system, especially for the detection of multi-scale targets and the real-time problem of detection. In the traffic sign detection process, the scale of the targets changes greatly, which will have a certain impact on the detection accuracy. Feature pyramid is widely used to solve this problem but it might break the feature consistency across different scales of traffic signs. Moreover, in practical application, it is difficult for common methods to improve the detection accuracy of multi-scale traffic signs while ensuring real-time detection. In this paper, we propose an improved feature pyramid model, named AF-FPN, which utilizes the adaptive attention module (AAM) and feature enhancement module (FEM) to reduce the information loss in the process of feature map generation and enhance the representation ability of the feature pyramid. We replaced the original feature pyramid network in YOLOv5 with AF-FPN, which improves the detection performance for multi-scale targets of the YOLOv5 network under the premise of ensuring real-time detection. Furthermore, a new automatic learning data augmentation method is proposed to enrich the dataset and improve the robustness of the model to make it more suitable for practical scenarios. Extensive experimental results on the Tsinghua-Tencent 100K (TT100K) dataset demonstrate that compared with several state-of-the-art methods, our method is more universal and superior.

## 1 Introduction

The traffic sign recognition system is the data foundation of intelligent transportation systems (ITS) and unmanned driving system, balancing the accuracy and real-time performance of the traffic sign detection and recognition technology, which plays an important role in the subsequent decision-making of ITS and unmanned driving system [1].

In recent years, most of the state-of-the-art object-detection algorithms have used convolutional neural networks (CNNs) and have achieved fruitful results in target detection tasks, such as the two-stage detectors Faster R-CNN [2], R-FCN [3], the one-stage detectors SSD [4], and YOLO [5]. However, directly applying these methods to traffic sign recognition is hard to achieve satisfactory results in practical application. The target recognition and detection of the vehicle-mounted mobile terminal require high accuracy for targets of different scales, and high requirements for recognition speed, which means to meet the two requirements of accuracy and real-time [6, 7].

Traditional CNNs use a large number of parameters and floating-point operations per second (FLOP) to achieve better detection performance. For example, VGG-16 [8] has about 138M parameters and requires 14.9B FLOPs to process an image of size 608×608. However, mobile devices (*e.g.* smartphones and self-driving cars) with limited memory and computation resources cannot be used for deployment and inference for larger networks. As a one-stage detector, the YOLOv5 [9] is used in this paper because of the advantages of low computation and fast recognition speed.

In this paper, an improved YOLOv5 network is proposed, which not only ensures that the model size can meet the requirements of deployment on the vehicle side but also improve the ability of multi-scale targets and meet the real-time requirement.

The main contributions of our work are summarized as follows:

- A novel feature pyramid network is proposed in this paper. Through adaptive feature fusion and receptive field enhancement, it retains the channel information in the feature transfer process to a large extent and learns different receptive fields in each feature map adaptively to enhance the representations of feature pyramids, effectively improving the accuracy of multi-scale targets recognition.
- A new automatic learning data augmentation strategy is proposed. Inspired by AutoAugment [10], the latest data augmentation operations have been added. The improved data augmentation method effectively improves the model training effect and the robustness of the training model, which has more practical significance.
- Unlike the existing YOLOv5 network, the current version is improved to reduce the impact of scale invariance. Meanwhile, it can be deployed on the mobile terminal of the vehicle to detect and recognize traffic signs in real-time.

The rest of this paper is organized as follows: Section 2 introduces related works about CNN-based traffic sign detection and data augmentation. Section 3 introduces the details of the proposed method to detect and recognize the traffic signs efficiently in real-time. The experimental results and analysis are presented in Section 4. Finally, the conclusion is described in Section 5.

## 2 Related Works

### 2.1 CNN-based traffic sign detection

At present, CNN as a popular algorithm for deep learning has a wide range of applications in computer vision, natural language processing, visual-semantic alignments, and other fields [11-14]. According to whether region proposal is required, it can be divided into two categories: single-stage detection and two-stage detection. Single-stage detection is often used in traffic detection due to its fast detection performance.

Shao *et al*. [15, 16] proposed an improved Faster R-CNN to traffic sign detection. They simplified the Gabor wavelet through a regional suggestion algorithm to improve the recognition speed of the network. Zhang *et al*. [17] modified the number of convolutional layers in the network based on YOLOv2, proposed an improved one-stage traffic sign detector, and used the China Traffic Sign Dataset for training to make it better adapted to Chinese traffic road scenes. A novel perceptual generative adversarial network was developed for the recognition of small-sized traffic signs [18], which boosts detection performance by generating super-resolve representations for small traffic signs. Aiming the scale variety problems in traffic sign detection, SADANet [19] combines a domain adaptive network with a multi-scale prediction network to improve the ability of the network to extract multi-scale features.

Most of the above-mentioned networks use single-stage detection and only use single-scale depth features, so it is difficult for them to have superior detection and recognition performance in sophisticated traffic scenes. Traffic sign instances of different scales have great differences in visual features, and the proportion of traffic signs in the entire traffic scene image is very small. Therefore, the scale variety problem has become a major challenge in traffic sign detection and recognition. And learning scale-invariant representation is critical for target recognition and location[20]. At present, this challenge is handled mainly from two aspects: network architecture modification and data augmentation [21].

At present, multi-scale features are widely used in high-level object recognition to improve the recognition performance of multi-scale targets [22]. Feature Pyramid Network (FPN) [23], as a commonly used multi-layer feature fusion method, uses its multi-scale expression ability to derive many networks with high detection accuracies, such as Mask R-CNN [24] and RetinaNet [25]. It is worth noting that the feature maps will suffer from information loss due to the reduced feature channels and only contain some less relevant context information in the feature maps of other levels.

Moreover, FPN pays too much attention to the extraction and optimization of low-level features. As the number of channels decreases, high-level features will lose a lot of information, resulting in a decrease in the detection accuracy of large-scale targets [22]. In response to this problem, a simple yet effective method named receptive field pyramid (RFP) [26] is proposed to enhance the representation ability of feature pyramids and drive the network to learn the optimal feature fusion pattern [14].

2.2 Data augmentation
Data augmentation has been widely utilized for network optimization and proven to be beneficial in vision tasks[2, 8, 27], which can improve the performance of CNN, prevent over-fitting [28], and is easy to implement [29].

Data augmentation methods could be roughly divided into color transformation (*e.g.*, noise, blur, contrast, and color casting) and geometric transformation (*e.g.*, rotation, random cropping, translation, and zoom) [30]. These augmentation operations artificially inflate the training dataset size by either data warping or oversampling. Lv *et al.* [31] proposed five data augmentation methods dedicated to face detection, including landmark perturbation and a synthesis method for four features of hairstyles, glasses, poses, illuminations. Nair *et al.* [32] performed geometric transformation and color transformation on the dataset. The training dataset is expanded and enriched by random crop and horizontal reflections and applying PCA on color space to change the intensity of the RGB channel. These frequently used methods just do simple transformations and cannot simulate the complex reality. Dwibedi *et al.* [33] improved detection performance with the cut-and-paste strategy. Furthermore, the method of annotated instance masks with a location probability map is utilized to augment the training dataset [34], which can effectively enhance the generalization of the dataset. YOLOv4 [13] and Stitcher [35] introduce mosaic inputs that contain rescaled sub-images, which are also used in YOLOv5. However, these data augmentation implementations are manually designed and the best augmentation strategies are dataset-specific.

The effect of data augmentation strategies is related to the characteristics of the dataset itself, so the focus of recent work has shifted to learning data augmentation strategies directly from the data itself. Tran *et al.* [36] generated augmented data, using the Bayesian approach, based on the distribution learned from the training set. Cubuk *et al.* [10] proposed a new data augmentation method that can automatically search for improved data augmentation policies, named AutoAugment.

## 3 Proposed Method
### 3.1 The improved YOLOv5s network framework
As the latest model in the current YOLO series, the superior flexibility of YOLOv5 makes it convenient for rapid deployment on the vehicle hardware side [37]. YOLOv5 contains four models, namely YOLOv5s, YOLOv5m, YOLOv5l, and YOLOv5x. YOLOv5s is the smallest model of the YOLO series and is more suitable for deployment on a vehicle-mounted mobile hardware platform due to its memory size of 14.10M, but the recognition accuracy cannot meet the requirements of accurate and efficient recognition, especially for the recognition of small-scale targets.

The basic framework YOLOv5 can be divided into four parts: Input, Backbone, Neck, and Prediction [37]. The Input part enriches the dataset with mosaic data augmentation, which has low requirements for hardware devices and low computational cost. However, it will cause the original small targets in the dataset to become smaller, resulting in the deterioration of the generalization performance of the model. The Backbone part is mainly composed of CSP modules, which perform feature extraction through the CSPDarknet53 [13]. FPN and Path Aggregation Network (PANet) [38] are used to aggregate the image feature at this stage in Neck. Finally, the network performs target prediction and output through the Prediction.

In this paper, the AF-FPN and the automatic learning data augmentation are introduced to solve the problem of incompatibility between model size and recognition accuracy, and further improve the recognition performance of the model. The original FPN structure is replaced by AF-FPN to improve the ability to recognize multi-scale targets and make an effective trade-off between recognition speed and accuracy [26]. Moreover, we remove the mosaic augmentation in the original network and use the best data augmentation methods according to the automatic learning data augmentation policy to enrich the dataset and improve the training effect. The improved YOLOv5s network structure is shown in Fig. 1.

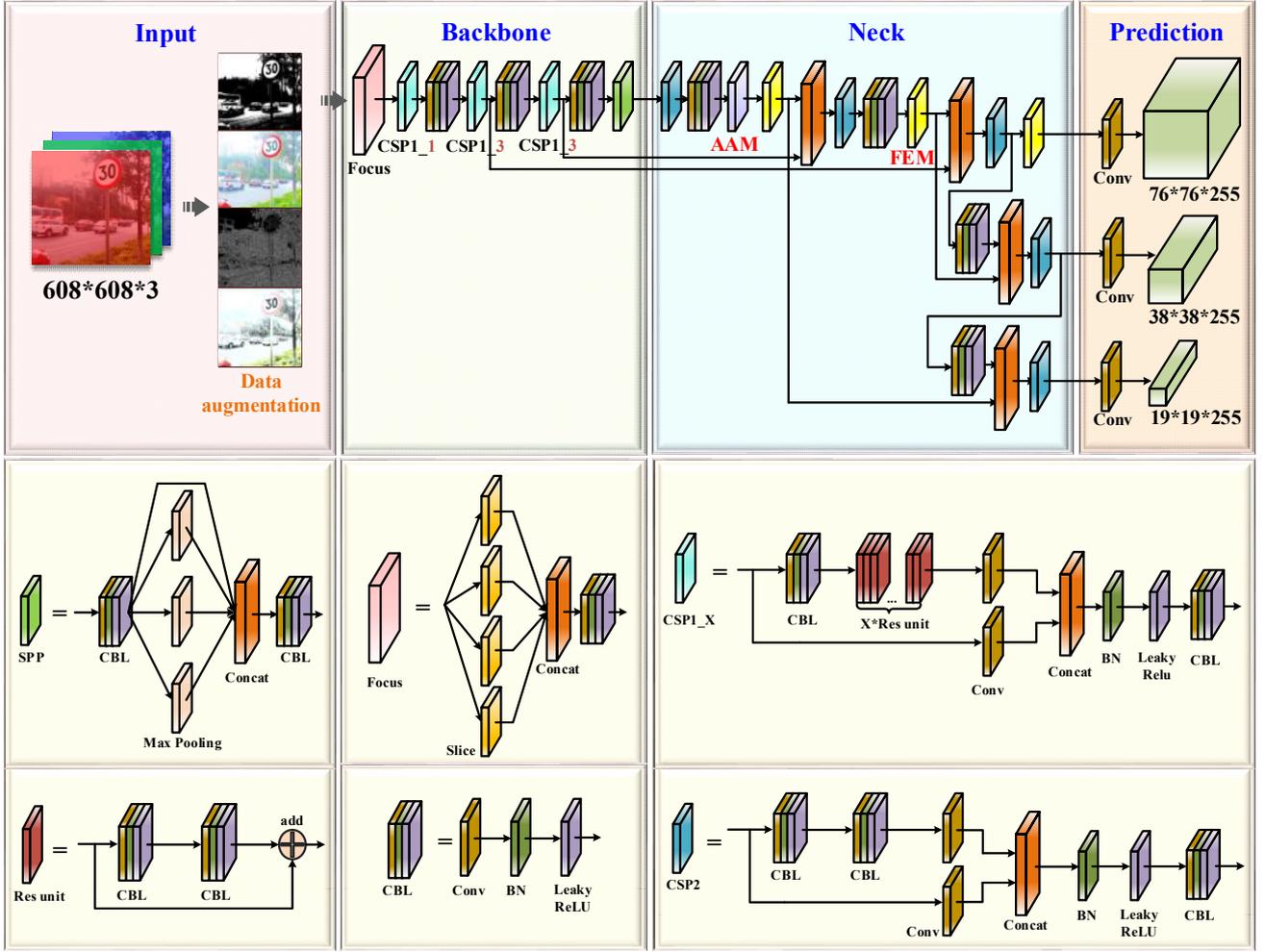

**Fig. 1** The architecture of the proposed YOLOv5s network

In Prediction, generalized IoU (GIoU) [39] loss is used as the loss function of the bounding box and the weighted non-maximum suppression (NMS) [40] method is used for NMS. The loss function is as follows:

$$L_{GIoU} = 1 - IoU + \frac{|C - (B \cup B^{gt})|}{|C|} \quad (1)$$

$$IoU = \frac{|A \cap B|}{|A \cup B|} \quad (2)$$

where $C$ is the smallest box covering $B$ and $B^{gt}$. $B^{gt}=(x^{gt}, y^{gt}, w^{gt}, h^{gt})$ is the ground-truth box, and $B=(x, y, w, h)$ is the predicted box.

However, when the predicted box is inside the ground-truth box and the size of the predicted box is the same, the relative positions of the predicted box and the ground-truth box cannot be distinguished.

In this paper, the GIoU is replaced by complete IoU (CIoU) [41] loss. Based on GIoU loss, the CIoU loss considers the overlap area, central point distance of bounding boxes, and the consistency of aspect ratios for bounding boxes. The loss function can be defined as:

$$R_{CIoU} = \frac{\rho^2(b, b^{gt})}{c^2} + \alpha v \quad (3)$$

$$v = \frac{4}{\pi^2}(arctan\frac{w^{gt}}{h^{gt}} - arctan\frac{w}{h})^2 \quad (4)$$

$$L_{CIoU} = 1 - IoU + R_{CIoU} \quad (5)$$

where $R_{CIoU}$ is the penalty term, which is defined by minimizing the normalized distance between central points of two bounding boxes. $b$ and $b^{gt}$ denote the central points of $B$ and $B^{gt}$, $\rho(\cdot)$ is the Euclidean distance, and $c$ is the diagonal length of the smallest enclosing box covering the two boxes. $\alpha$ is a positive trade-off parameter, and $v$ measures the consistency of the aspect ratio.

And the trade-off parameter $\alpha$ is defined as

$$\alpha = \frac{v}{(1-IoU)+v} \quad (6)$$

which the overlap area factor is given higher priority for regression, especially for non-overlapping cases.

### 3.2 AF-FPN structure

Based on the traditional feature pyramid network, AF-FPN adds the adaptive attention module (AAM) and the feature enhancement module (FEM). The former part reduces the loss of context information in the high-level feature map due to the reduced feature channels. The latter part enhances the representation of feature pyramids and accelerates the inference speed while achieving state-of-the-art performance. The structure of the AF-FPN is shown in Fig. 2.

The input image generates feature maps {$C_1$, $C_2$, $C_3$, $C_4$, $C_5$} through multiple convolutions. $C_5$ generates the feature map $M_6$ through AAM. And $M_6$ is combined with $M_5$ by summation and propagated to fuse with other features at lower levels through a top-down path, and the receptive field is expanded through FEM after each fusion. PANet shortens the information path between lower layers and the topmost feature.

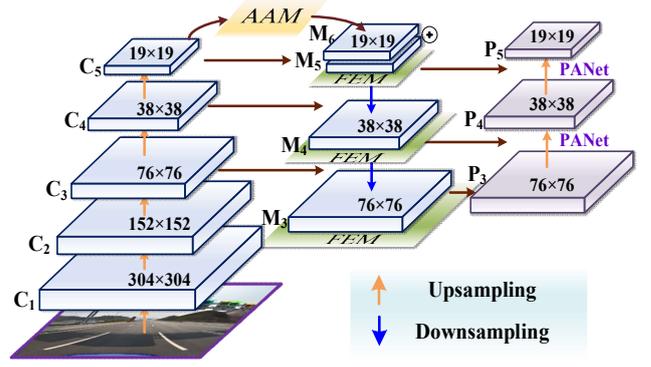

**Fig. 2** The architecture of the AF-FPN

The operation of the adaptive attention module can be performed in two steps. First of all, the multiple context features with different scales are obtained through the adaptive average pooling layer. The pooling coefficient $\beta$ is [0.1, 0.5], and it adaptively changes according to the target size in the dataset. Secondly, a spatial weight map is generated for each feature map through the spatial attention mechanism. Through the weight map, context features are fused to generate a new feature map, which contains multi-scale context information. The new feature map is combined with the original high-level feature map and propagated to fuse with other features at lower levels.

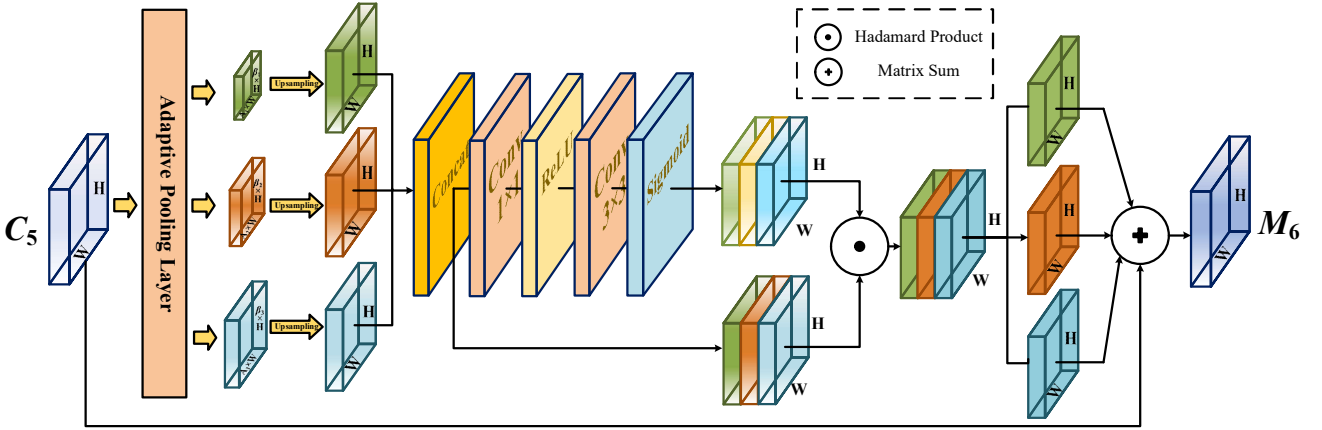

**Fig. 3** The architecture of the AAM

The specific structure of the AAM is shown in Fig. 3. As the input of the adaptive attention module, the size of $C_5$ is $S=h\times w$. It first obtains context features with different scales of ($\beta_1\times S$, $\beta_2\times S$, $\beta_3\times S$) through the adaptive pooling layer. Then each context feature undergoes a 1×1 convolution to obtain the same channel dimension 256. Bilinear interpolation is used to upsample them to the scale of $S$ for subsequent fusion. The spatial attention mechanism merges the channels of the three context features through a Concat layer, and then the feature map sequentially passes 1×1 convolution layer, ReLU activation layer, 3×3 convolution layer, and sigmoid activation layer to generate corresponding spatial weights for each feature map. The generated weight map and the feature map after the merged channel are subjected to the Hadamard product operation, which is separated and added to the input feature map $M_5$ to aggregate context features into

$M_6$. The final feature map has rich multi-scale context information, which to a certain extent alleviates the loss of information due to the reduction of the number of channels.

FEM mainly uses the dilated convolution to learn the different receptive fields in each feature map adaptively based on the varying scales of detected traffic signs, thereby improving the accuracy of multi-scale target detection and recognition. As shown in Fig. 4, it can be divided into two components: the multi-branch convolution layer and the branch pooling layer. The multi-branch convolution layer is used to provide different sizes of receptive fields for the input feature map through the dilated convolution. And the average pooling layer is used to fuse the traffic information from the three branch receptive fields to improve the accuracy of multi-scale prediction.

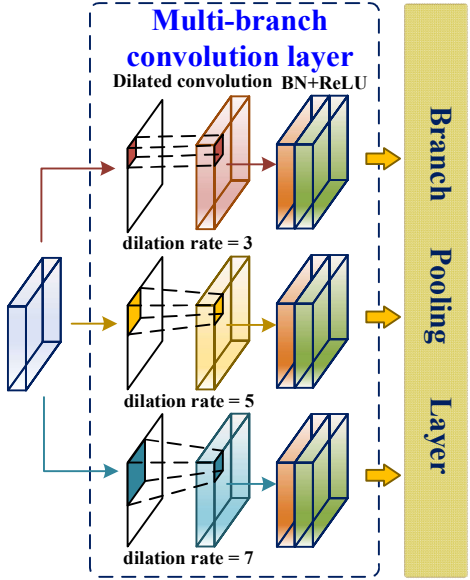

**Fig. 4** The architecture of the FEM

The multi-branch convolution layer consists of dilated convolution, BN layer, and ReLU activation layer. The dilated convolutions in the three parallel branches have the same kernel size but different dilation rates. Specifically, the kernel of each dilated convolution is 3×3 and the dilation rates $d$ is 1, 3, and 5 for different branches.

Dilated convolutions support exponentially expanding receptive fields without losing resolution or coverage [42]. However, in the convolution operation of dilated convolution, the elements of the convolution kernel are spaced, and the size of the space depends on the dilation rates, which is different from the elements of the convolution kernel that are all adjacent in the standard convolution operation.

The convolution kernel changed from 3×3 to 7×7 and the receptive field of this layer is 7×7. The formula for the receptive field of dilated convolution is as follows:

$$r_1 = d \times (k-1) + 1 \qquad (7)$$

$$r_n = d \times (k-1) + r_{n-1} \qquad (8)$$

where $k$ and $r_i$ denote the kernel size and dilation rate, respectively. And $d$ denote the stride of the convolution.

The branch pooling layer [43] is proposed to fuse information from different parallel branches and avoid introducing additional parameters. The averaging operation is utilized to balance the representation of different parallels branches during training, which enables a single branch to implement inference during the test. The expression is as follows:

$$y_p = \frac{1}{B} \sum_{i=1}^{B} y_i \qquad (9)$$

where $y_p$ denotes the output of the branch pooling layer. $B$ represents the number of parallel branches and we set $B = 3$.

### 3.3 Data Augmentation

The augmentation policy consists of two parts: search space and search algorithm [44]. The search space contains 5 sub-strategies, each of which consists of two simple image enhancement operations applied in sequence. One of the sub-policies are be chosen at random and applied to the current image. In addition, each operation is also associated with two hyperparameters: the probability of applying the operation and the magnitude of the operation [10]. The operations we used in the experiment include the latest data augmentation methods such as Mosaic [13], SnapMix [45], Erasing, CutMix, Mixup, and Translate X/Y. In total, we have 15 operations in our search space. Each operation also comes with a default range of magnitude. We discretize the range of magnitude into $D=11$ values that follow the uniform distribution so that we can use a discrete search algorithm to find them. Similarly, we also discretize the probability of applying one of all operation into $P=10$ values (also following a uniform distribution). Finding each sub-policy becomes a search problem in a space of $(19 \times D \times P)^2$ possibilities. Therefore, the search space with 5 sub-policies then has roughly $(19×D×P)^{2×5}$ possibilities and requires an efficient search algorithm to navigate this space [46]. Fig. 5 shows the policy with 5 sub-policies in the search space.

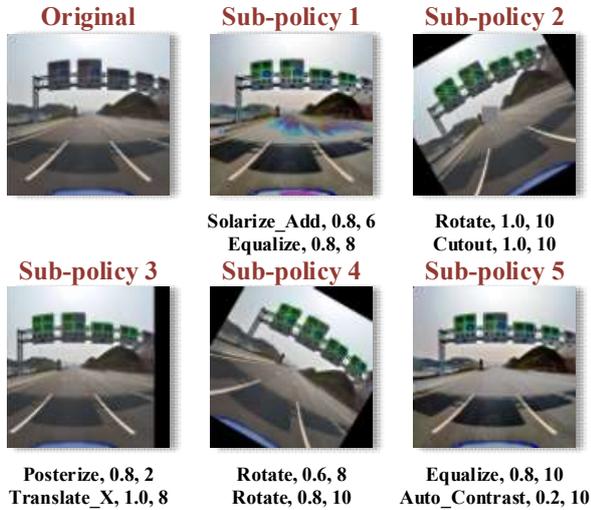

**Fig. 5** An example of a policy with 5 sub-policies

Through the search space, the problem of searching for a learned augmentation policy into a discrete optimization problem. Reinforcement Learning [47] is used as the search algorithm, which has two components: a controller RNN and the training algorithm. The controller RNN is a recurrent neural network, and the proximal policy optimization (PPO) [46] with a learning rate of 0.00035 is used as the network training algorithm. The controller RNN predicts a decision produced by a softmax at each step and the prediction is then fed into the next step as an embedding from the search space. Totally, the controller has 30 softmax predictions to predict 5 sub-policies, each of which has two operations, and each operation requires an operation type, magnitude, and probability. We applied the automatic learning data augmentation method to the TT100K dataset, and then used the best data augmentation policy obtained through training.

## 4 Experiments and Analysis

In this section, we comprehensively evaluate the improved YOLOv5 model through the TT100K [48, 49] dataset, which includes 182 types of traffic signs instances with detailed annotations and covers different actual traffic environments. And about 42.5% of traffic signs in TT100K are small objects, which means it is more suitable for actual vehicle-mounted target recognition. The size distribution of traffic sign instances in the dataset is displayed in Fig. 6.

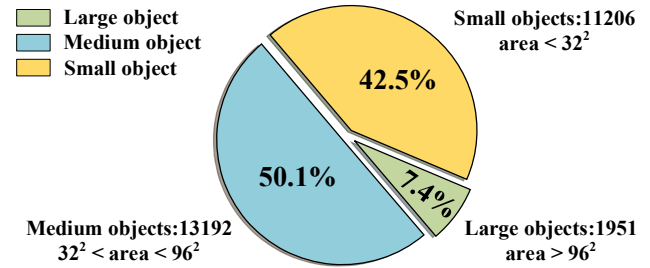

**Fig. 6** Size distribution of sign instances from the TT100K

### 4.1 Experimental setting

Considering the fixed size of input demanded by the YOLOv5 network, we resized the images to uniform dimensions of 608×608. The training and validation datasets include 9146 images and the test dataset include 1121 image from TT100K. In the process of training, the initial value of the learning rate was 0.01, and use the cosine annealing strategy to reduce the learning rate. The epochs and the batch size are set to 500 and 32, respectively. Our experiments were performed on a Linux4.15.0-142-generic Ubuntu 18.04 with Intel(R) Xeon(R) Silver 4210R CPU @ 2.40GH, 8×32GB DDR4 and 8×TITAN Xp, 12GB memory. The mobile device used in the experiment is Jetson Xavier NX, and an external USB3.0 industrial camera.

### 4.2 Experimental analysis

To demonstrate the advantages of the proposed method in traffic sign detection, we evaluated our method on TT100K and compared it with the original YOLOv5, YOLOv5-Lite [50], Efficientdet [51], YOLOv5-face [52], M2det [53], SSD, and YOLOv3 [54]. We evaluated performance using metrics including model size, parameters, floating-point operations per second (FLOPs), mean average precision (mAP), average precision of large, medium, and small size targets ($AP_L$, $AP_M$, $AP_S$), and frames per second (FPS). The specific results are shown in Table I.

Table I

Comparison of our method with other method on the TT100K dataset

| Method | Model | Parameters | FLOPs | $AP_S$ | $AP_M$ | $AP_L$ | mAP@0.5 | FPS |
|---|---|---|---|---|---|---|---|---|
| Efficientdet-d0 | 15.15M | 3.752M | 2.50B | - | 0.3980 | 0.5526 | 0.5786 | 26 |
| M2det | 340M | 147M | 16.35G | 0.0300 | 0.3198 | **0.6586** | 0.4658 | 12 |
| Mobilenet-SSD | 118M | 25.067M | 29.20G | 0.0275 | 0.2443 | 0.5261 | 0.3200 | 22 |
| YOLOv3 | 119.6M | 59.578M | 158.00G | 0.3889 | 0.4454 | 0.4288 | 0.6186 | 27 |
| YOLOX-s | 69.0M | 9.010M | 27.03G | 0.4122 | **0.5975** | 0.6081 | **0.6860** | 59 |
| YOLOv5-Lite-g | 11.6M | 5.277M | 15.60G | 0.3218 | 0.4914 | 0.4957 | 0.5431 | 71 |
| YOLOv5 | 14.6M | 7.193M | 17.90G | 0.3567 | 0.5142 | 0.5184 | 0.6018 | **105** |
| Ours | 16.3M | 8.039M | 17.90G | **0.4146** | 0.5783 | 0.5817 | 0.6514 | 95 |

Firstly, it can be seen that the model size of the proposed method is 16.3M, which is easy to be deployed on a mobile platform so that it can be used for real-time shooting and recognition on the vehicle side. The amount of parameters in the training process is slightly higher than that of Efficientdet and YOLOX. FLOPs is 17.9G. which is only 3.3G larger than the optimal YOLOv5-Lite. It can be seen from these two indicators that our method has a faster training speed and requires less hardware equipment, which is convenient for popularization. Excessive reduction of the number of parameters and calculations will lead to a decrease in the detection effect of the final training model. Secondly, our method achieves the mAP of 65.14% on all 182 traffic sign classes, which is second only to YOLOX. Although the $AP_L$ of our method is lower than M2det and YOLOX, the recognition accuracy on small targets is 41,46%, which is significantly higher than other methods. Finally, FPS is used as a metric to evaluate the speed of target detection, indicating that our method can meet the real-time requirements of detection on the mobile terminal. In general, our method has high accuracy for multi-scale target detection and can achieve a balance between recognition accuracy and recognition speed. The model size is suitable for deployment on the mobile terminal and has practical application significance.

In addition, traffic sign detection is a multi-category and multi-target recognition task, and the false detection rate and missed detection rate are also important metrics to measure the detection network. In order to verify the missed detection requirements of the proposed method in real-time traffic sign detection, Log-Average Miss RATE (LAMR) [55] is selected as the evaluation index. LAMR reflects the relationship between the false negative (FP) of each image and the missed detection rate. The lower the FP, the better the detection performance of traffic signs. We selected the top 19 traffic sign categories in the dataset, and compared the missed detections of each method on these types of traffic signs, as shown in Fig. 7. It can be seen that the missed detection rate of our method for traffic sign recognition is significantly lower than other methods, and it has practical application significance. However, the missed detection rate of several types of traffic signs such as ip, $w_{57}$, and po is still high, and we will make further improvements in future research.

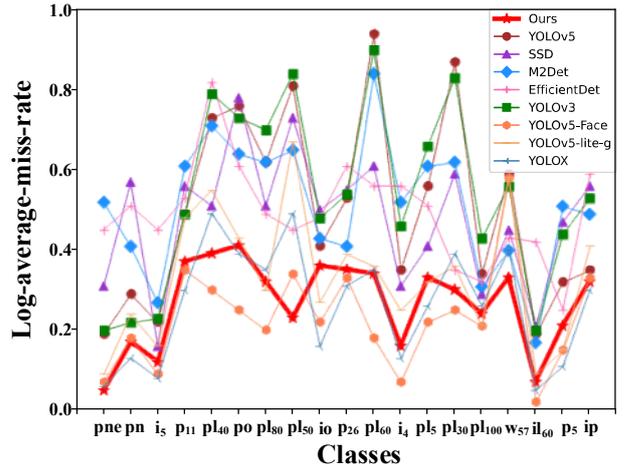

**Fig. 7** Comparison of the miss detection rate of each method on 19 types of traffic signs

We visualize the method proposed in this paper, as shown in Fig. 8(a). It can be observed that our method successfully recognized the multi-sized traffic signs on the actual traffic scene with high recognition accuracy, and there are almost no missed detections and false detections. Meanwhile, we transplanted the trained model to the mobile platform and connected an external camera to shoot the actual road scene. Real-time traffic sign detection and recognition are performed on the captured images, and the recognition results are displayed on the LED screen, as shown in Fig. 8(b).

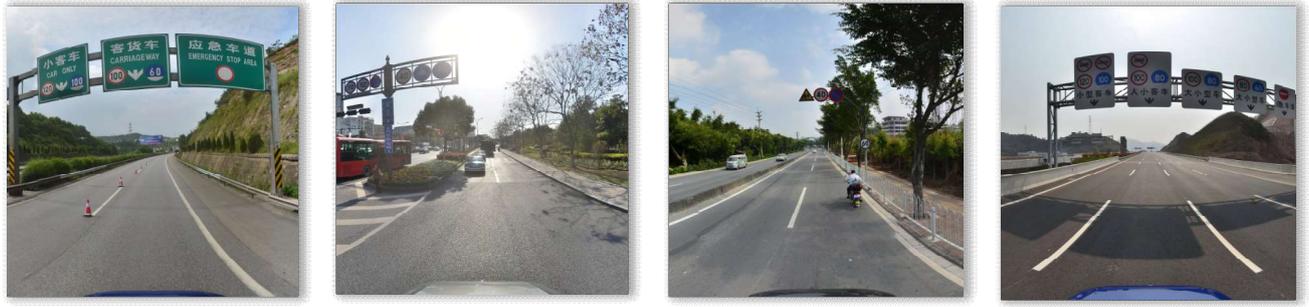

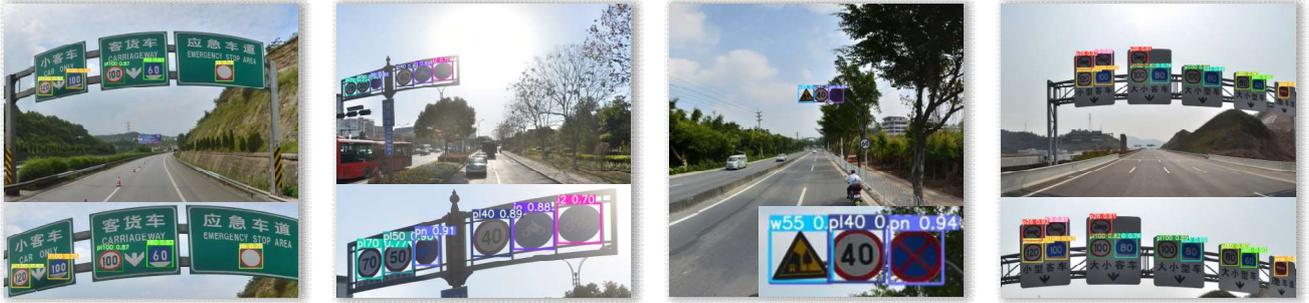

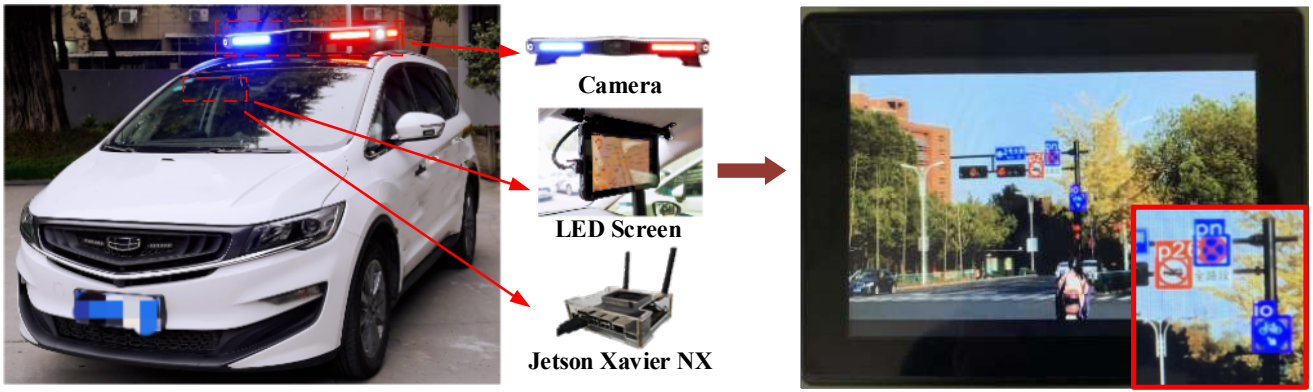

**Fig. 8** (a) Some examples were detected by our method on the TT100K dataset; (b) The mobile device deployment and the detection example of shooting through the camera.

4.3 Ablation study

To more intuitively demonstrate the better performance of the proposed method for traffic sign detection and recognition, we conduct the ablation study, and the results are shown in Table II.

Table II
Ablation study

| Method | Model | Param | FLOPs | FPS | mAP |
|---|---|---|---|---|---|
| YOLOv5s | 14.6M | 7.193M | 17.9G | 105 | 0.6018 |
| YOLOv5s+Aug | 16.3M | 7.193M | 17.9G | 105 | 0.6131 |
| YOLOv5s+AF-FPN | 14.6M | 8.039M | 17.9G | 95 | 0.6267 |
| Ours | 16.3M | 8.039M | 17.9G | 95 | **0.6514** |

Table II shows the ablation result of incrementally adding the components training on the YOLOv5s model. As observed from the results, the standard YOLOv5s provides a detection mAP of 60.18%. Integrating the data augmentation and the AF-FPN improves the mAP to 61.31% and 62.67%, respectively. The mAP of our method on the TT100K dataset is 4.96% higher than that of the YOLOv5s, which means the proposed method achieves impressive performance in target and recognition. At the same time, the model size and parameters amount only slightly increase, and the FLOPs do not change, which means that the training speed of the improved network and the requirements for training equipment are basically unchanged. These ensure that our method can be easily deployed on the vehicle side. Although FPS decreases by 10, it still meets the requirements of real-time detection on the vehicle side.

## 5 Conclusion

In this paper, we proposed a real-time traffic sign detection network based on modified YOLOv5s, which achieves better detection performance than state-of-art one-stage detectors. In this work, the proposed AF-FPN structure improves the information extraction ability of feature maps and its representation ability for detecting multi-scale objects. And the new data augmentation strategy enriches the traffic sign dataset by adding Noise, Mosaic, and other methods to improve the training effect of the model. The empirical results verified that the proposed method could achieve state-of-the-art performance with a fast inference speed, the detection speed on the vehicle side is 95 FPS. The proposed method provides the input feature map of different receptive fields and fuses the receptive field pyramids for the target traffic signs. Therefore, the improved network can enhance the recognition accuracy of multi-scale targets without introducing additional calculations, the mAP has increased by 4.96% compared to the original network on the TT100K. Due to the size of the trained model being small, it is easy to deploy on the mobile device of the vehicle and perform real-time recognition and detection of the road scene. However, in practical applications, a faster vehicle speed will cause the motion blur of the image, which will affect the recognition result. In the future, we plan to explore a better performance detection model for high-speed moving targets.

**Acknowledgments** This work was supported in part by Zhejiang Provincial Key Lab of Equipment Electronics under Grant 2019E10009; in part by the Key Research and Development Program of Zhejiang Province under Grant 2020C01110.